\documentclass[sigconf]{acmart}

\AtBeginDocument{%
}

\copyrightyear{2024}
\acmYear{2024}
\setcopyright{rightsretained}
\acmConference[HRI '24 Companion]{Companion of the 2024 ACM/IEEE International Conference on Human-Robot Interaction}{March 11--14, 2024}{Boulder, CO, USA}
\acmBooktitle{Companion of the 2024 ACM/IEEE International Conference on Human-Robot Interaction (HRI '24 Companion), March 11--14, 2024, Boulder, CO, USA}
\acmDOI{10.1145/3610978.3640583}
\acmISBN{979-8-4007-0323-2/24/03}

\makeatletter
\gdef\@copyrightpermission{
  \begin{minipage}{0.3\columnwidth}
   \href{https://creativecommons.org/licenses/by/4.0/}{\includegraphics[width=0.90\textwidth]{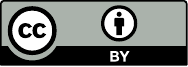}}
  \end{minipage}\hfill
  \begin{minipage}{0.7\columnwidth}
   \href{https://creativecommons.org/licenses/by/4.0/}{This work is licensed under a Creative Commons Attribution International 4.0 License.}
  \end{minipage}
  \vspace{5pt}
}
\makeatother

\begin{document}

\title{\textbf{Developing Autonomous Robot-Mediated Behavior Coaching Sessions with Haru}
}


\author{Matouš Jelínek}
\email{matous@sdu.dk}
\orcid{0000-0002-5303-9238}
\affiliation{%
 \institution{University of Southern Denmark}
  \city{Sønderborg}
  \country{Denmark}
  \postcode{6400}
}
\author{Eric Nichols}
\email{eric.nichols@ieee.org}
\orcid{0000-0003-0734-6621}
\affiliation{%
  \institution{Honda Research Institute Japan}
  \streetaddress{8-1 Honcho}
  \city{Wako}
  \country{Japan}
}

\author{Randy Gomez}
\email{r.gomez@jp.honda-ri.com}
\orcid{0000-0002-3191-6818}
\affiliation{%
  \institution{Honda Research Institute Japan}
  \city{Wako}
  \country{Japan}
}

\renewcommand{\shortauthors}{Matouš Jelínek, Eric Nichols, \& Randy Gomez}

\begin{abstract}
This study presents an empirical investigation into the design and impact of autonomous dialogues in human-robot interaction for behavior change coaching. We focus on the use of Haru, a tabletop social robot, and explore the implementation of the Tiny Habits method \cite{fogg_tiny_2019} for fostering positive behavior change. The core of our study lies in developing a fully autonomous dialogue system that maximizes Haru's emotional expressiveness and unique personality. 
Our methodology involved iterative design and extensive testing of the dialogue system, ensuring it effectively embodied the principles of the Tiny Habits method while also incorporating strategies for trust-raising and trust-dampening. 
The effectiveness of the final version of the dialogue was evaluated in an experimental study with human participants (N=12). The results indicated a significant improvement in perceptions of Haru's liveliness, interactivity, and neutrality. Additionally, our study contributes to the broader understanding of dialogue design in social robotics, offering practical insights for future developments in the field. 
\end{abstract}





\maketitle

\vspace*{-2mm}
\section{Introduction}
The use of social robots in behavior change coaching presents a unique intersection of human-robot interaction (HRI), psychology, and AI. Social robots have been previously utilized in behavior change coaching, with various studies underscoring their potential in this field \cite{bodala_teleoperated_2021,arendsen_use_2010,jelinek_role_2021}. Social robots offer new ways of interaction, learning, and behavior adaptability that can complement traditional human-led coaching methods. 

Trust emerges as a fundamental element in the realm of HRI, particularly within the scope of behavior change coaching. The concept of trust calibration, which involves aligning a human's trust level with a robot's actual capabilities \cite{lee_trust_2004,hancock_meta-analysis_2011}, plays a pivotal role. Proper calibration of trust is crucial for accurately reflecting a robot's competence and reliability, thereby preventing over-reliance or under-utilization of robotic systems. This calibration is integral for effective and safe interactions \cite{de_visser_towards_2020}.

Previous work suggests that one way of raising and maintaining the level of trust in HRI is the robot's emotional expressions, encompassing aspects such as emotional routines, voice modulation, and timing. These elements significantly influence the robot's perceived trustworthiness and level of engagement \cite{breazeal_emotion_2003,tielman_adaptive_2014,nass_does_2001}. 

In behavior change coaching, where personal and emotional interactions are paramount \cite{fogg_tiny_2019}, the rich expressivity of a robot can profoundly impact the coaching effectiveness. The importance of emotional expressions in enhancing the coaching experience was demonstrated, for example in \cite{jelinek_role_2021-1}, utilizing the Tiny Habit method and the social robot Haru. 

The study revealed that emotional expressions in Haru significantly enhanced the perceived quality of the lesson and the retention of the habit practiced. Additionally, Haru's presence, particularly when using emotional behaviors, led to higher confidence in participants regarding their behavior change and increased the likelihood of habit retention compared to control groups presented with the same information via web content \cite{jelinek_role_2021}. 

However, this study had various limitations: they relied on a wizard operator and used only virtual simulation of the robot. These constraints potentially limited the depth of interaction and the naturalness of the robot's behavior.

Our study addresses these limitations by taking a novel approach. We developed a new fully-autonomous dialogue for Haru, aiming to harness the full emotional potential of the physical robot, allowing for more organic and realistic interactions, critical in the context of behavior change coaching. By integrating emotional routines and new voice design \cite{Nichols_voice} into Haru's autonomous dialogue, we aim to deepen the level of engagement and trust in HRI.

Moreover, our research extends the understanding of how emotionally intelligent interactions influence users' perception of the robot. In doing so, we seek to bridge the gap between the theoretical potential of social robots in behavior change coaching and their practical application in real-world scenarios.

\vspace*{-2mm}
\section{\textbf{Method}}

In developing the new dialogue for behavior change coaching, we embraced the Design Thinking method \cite{noauthor_what_nodate}. This method is renowned for its hands-on, user-centric approach to problem-solving.

Our primary goal was to craft a dialogue that leverages the Tiny Habits method, explicitly tailored for Haru, a social robot. We focused on utilizing Haru's unique capabilities to enhance the effectiveness of the dialogue in behavior change coaching. This section will delve into the development journey of the dialogue and highlight the key design decisions made throughout this process.

\vspace*{-2mm}
\subsection{\textbf{Behavior change coaching} }
We selected The Tiny Habit Method \cite{fogg_tiny_2019} as a framework for behavior change coaching. The Tiny Habits method is grounded in several theoretical principles from behavior change science - e.g., \cite{ryan2000self,ajzen1991theory,prochaska1982transtheoretical}. It is based on Fogg's Behavioral model \cite{fogg_behavior_2009}, which states that behavior is a product of three elements: motivation, ability, and prompt. For behavior to occur, a person must be \textit{i)} sufficiently motivated, \textit{ii)} has the ability to perform the behavior, and \textit{iii)} have the proper trigger has to be present. The Tiny Habits method focuses on addressing all those aspects. It involves anchoring new, small habits to existing routines, simplifying aspirations into manageable actions, and rewarding completion with positive emotions to integrate these habits into daily life.

\vspace*{-2mm}
\subsection{Social Robot}
We utilized Haru, an experimental tabletop social robot, as our target robot. Haru's design, inspired by animated characters \cite{gomez_haru_2018}, features two eyes with LCD screens and moving rims equipped with addressable LED strips alongside an LED matrix mouth. Haru boasts seven degrees of freedom, including eye tilt, rotation, inner eye movement, base rotation, and body leaning, allowing it to create dynamic emotional signals \cite{gomez_haru_2018,gomez_haru_emotion_2020}.
Haru's design emphasizes the upper and middle body for expressivity, based on studies showing these areas are crucial in conveying emotions \cite{gomez_haru_emotion_2020}. Without hands, Haru's emotional expressions are conveyed through eye and body movements, similar to the emotional affordances of hands and arms.

Haru's expressivity is enhanced by its custom Text-to-Speech (TTS) voice, featuring a range of vocal genres for emotive communication. The voice includes seven vocal genres: default, cheeky, high-energy, question, sad, serious, and whiny, for a broad emotional range \cite{Nichols_voice}. Haru's library of over 100 routines also supports multimodal expressions, including body and eye motions, animations, and sounds, covering a full spectrum of emotions \cite{gomez_haru_emotion_2020}.

The robot is using Google Speech-to-Text API for automatic speech recognition (ASR), and Intent Classification and Entity Recognition Models developed by Honda Research Institute. The dialogue structure is then loaded into behavior trees,  with all the components interacting with the robot hardware using ROS \cite{nichols_design_2021}.  This system, which combines off-the-shelf components with custom robot elements, was  previously tested in other applications \cite{nichols2022hey}, such as small talk scenarios and hospital dialogues.

\vspace*{-2mm}
\subsection{Implementation Process}
Our development, guided by Design Thinking, involved iterative refinement of the dialogue. Initially, we assessed the existing dialogue \cite{jelinek_role_2021}, identifying key issues to be changed, such as the verbosity of the dialogue, focus only on yes/no questions, or the need for a human operator to make all the decisions in the dialogue. Our primary goals were defined as to develop an autonomous dialogue aligned with Haru's personality \cite{Nichols_voice}, incorporating trust cues and enhancing emotional expressions, which we then ideated and converted into a first dialogue prototype. 
Pilot testing, conducted over six distinct iterations with volunteers, primarily engineers without previous knowledge of the dialogue, was instrumental in evolving the prototype into the final dialogue. The process focused on enhancing natural emotional responses and trust-building strategies in the dialogue flow.

We first describe the the new dialogue developed. Subsequently, we'll delve into specific implemented refinements, highlighting their potential applicability in other social robotics dialogues.

\vspace*{-2mm}
\subsection{New Dialogue Development}
The final version of the dialogue, in which Haru coaches participants to adopt a Tiny Habit, is divided into four main sections. In the first section (S1), Haru explains the importance of finding the right aspiration for behavior change and then provides suggestions on how to transform this aspiration into a small daily routine - the Tiny Habit. In the second section (S2), the participant and the robot collaboratively identify a routine that can serve as a reminder to conduct the new habit, referred to as the anchor moment. The third section (S3) focuses on reinforcing the newly developed Tiny Habit by celebrating the participant's successful completion of the new habit. In the last section (S4), Haru summarizes the theory and offers additional help before concluding the session.

\begin{figure}

    \centering
    \includegraphics[width=1\linewidth]{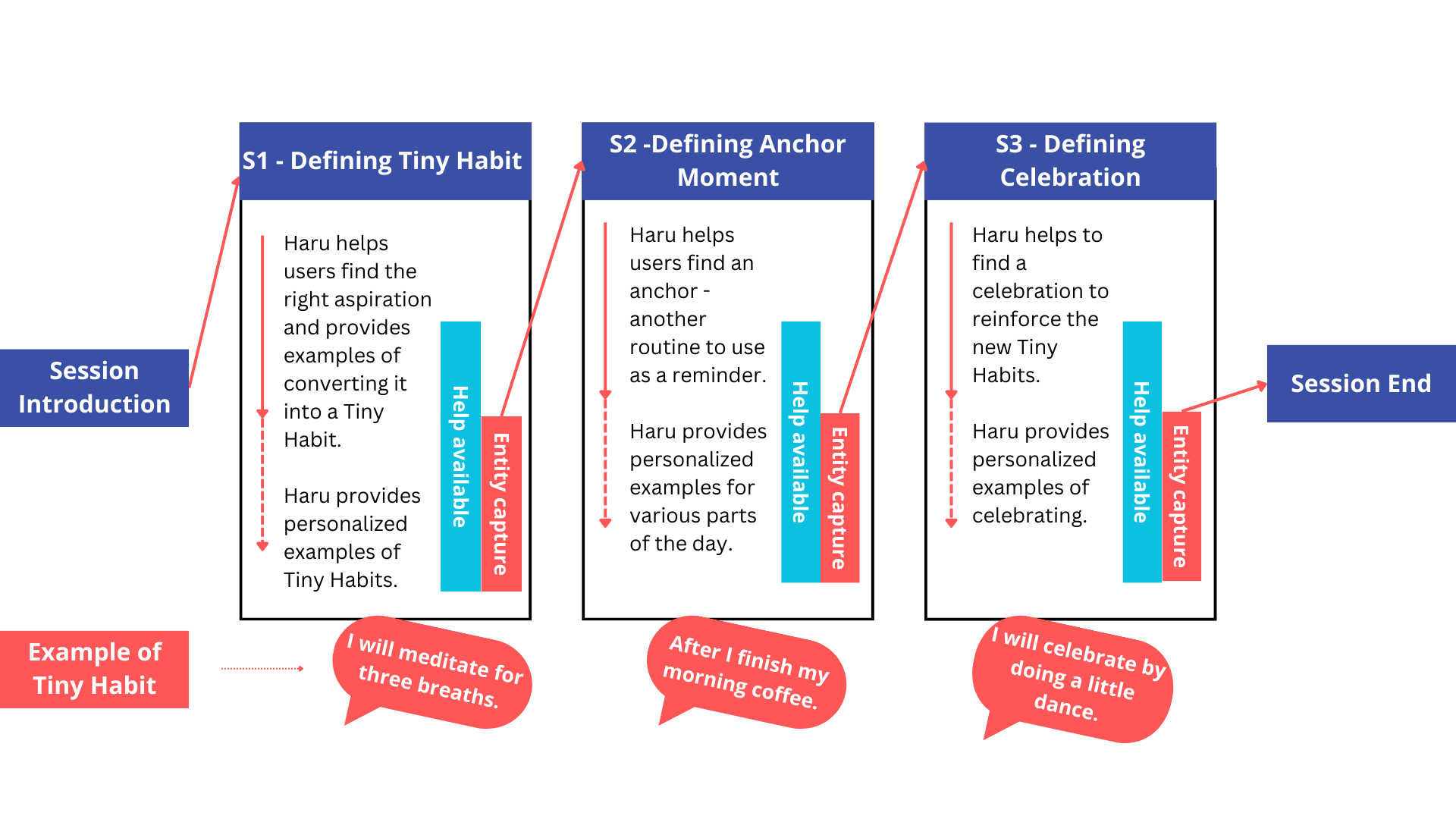}
    \vspace{-7mm} 
    \caption{Overview of the Dialogue Structure}
    \Description{The image is a diagram of a three-step session for defining and implementing Tiny Habits. The first step is "S1 - Defining Tiny Habit," where 'Haru' assists users in identifying a small, achievable habit tied to a larger aspiration and offers personalized examples. The second step is "S2 -Defining Anchor Moment," where 'Haru' aids in finding a routine to serve as a reminder for the habit. The third step is "S3 - Defining Celebration," where 'Haru' helps choose a celebration to reinforce the new habit. The diagram includes examples for each step, such as "I will meditate for three breaths" after the session introduction, "After I finish my morning coffee" as an anchor moment, and "I will celebrate by doing a little dance" to mark the celebration. Each step has a color-coded bar on the side indicating "Help available" and "Entity capture." The session begins with an introduction and ends with a celebration.}
    \vspace{-5mm} 
    \label{fig:dialog-structure}

\end{figure}

In the first three sections (S1-S3), the dialogue follows a consistent structure. In the first part (P1), which users can skip, Haru presents an example of how he addressed the material covered in the section in his own life. In the second part (P2), participants are asked how they want to address the section's topic - defining the Tiny Habit, finding the right anchor moment, or the celebration (e.g., ``What’s your aspiration you'd like to work on?"). Based on their response, the robot selects the most appropriate answer from one of the pre-defined categories (e.g., User reply: ``I want to start running"; Haru’s response: “Awesome! A dash of exercise can spark a fitness firework! One way to get into shape is by scaling down to doing two pushups or putting on your running shoes to get ready.”).

In the third part (P3), the robot confirms if the subjects have understood the concept and offers additional coaching tips. Simultaneously, the robot begins capturing potential user responses. The last part (P4) focuses on capturing the user's progress (e.g., their new Tiny Habit). Haru tries to capture the entity multiple times, offers additional guidance and examples of good practice, and provides a default entity for adoption if participants are unsure (e.g., ``Here's one last idea for your possible change. You seem a tad stressed. How about focusing on relaxation? Your new Tiny Habit could be, ‘I will take three mindful breaths.’ Does that sound like something you might use in the future?"). If they do not confirm, the dialogue concludes preliminarily.

If the entity is captured during the third and fourth parts of the dialogue, Haru confirms with the user whether it was captured correctly. If the answer is positive, the dialogue proceeds to the next section without the possibility of returning. 

\vspace*{-2mm}
\subsection{Dialogue Refinement}
\subsubsection{Emotional and Vocal Expressivity
}
We leveraged Haru's extensive library of emotional routines and a diverse range of voice genres to create a more engaging and human-like interaction. Our decision to implement these routines was grounded in theoretical principles and insights from pilot testing, which revealed a preference for routines that included sound and were shorter than 3.5 seconds, particularly crucial in extended dialogue segments.

Haru's emotional routines encompass the full spectrum of Ekman's basic emotions \cite{ekman1984expression}, with multiple variants for each emotional state. We strategically assigned these routines to specific parts of the dialogue, ensuring they did not disrupt the flow, particularly at the end of utterances or after questions. This approach aligns with affective grounding theory, emphasizing shared emotional understanding as vital in human-robot interaction \cite{jung_affective_2017}.

Other key enhancements in our dialogue design included integration of nuanced emotional expressions, such as back-channels like ``Oh!'' and nods, to indicate understanding and engagement. Furthermore, we tried to utilize emotional mimicry, adapting to the emotional context of interactions, informed by research on human emotional response patterns \cite{kraut_social_1979}.

These features are rooted in the concept that robots can express emotions as signals to reveal internal states, based on basic emotion theories. Our methodology also incorporated affective storytelling, where Haru's dialogue reflected appropriate emotional intonations corresponding to the narrative, further enhancing the interaction's intuitiveness and effectiveness \cite{chella_emotional_nodate}, and trustworthiness \cite{fischer_emotion_2019}.

\subsubsection{Empathy}
Our study emphasized empathy in the dialogue design to foster a more sympathetic and relatable interaction. Key to this was Haru's ability to respond to detected problems in the interaction with comforting statements such as ``Don't worry, I am here to help you,'' ``Take your time!" and ``No problem. Let's move forward!". This approach is rooted in theories of empathetic design \cite{picard_affective_1997,paiva_emotion_nodate,maldonado_we_2005}, highlighting the importance of empathy in effective interaction design. By incorporating these empathetic responses, Haru could better connect with users, enhancing the overall trust and effectiveness of the behavior change coaching process.

\subsubsection{Situation Awareness}
The concept of situation awareness in human-computer interaction underscores the importance of context for creating meaningful and trustworthy interactions \cite{lee_trust_2004,abowd_towards_1999}. In social human-robot interaction, a robot's situation awareness can significantly affect the robot's persuasive abilities \cite{fischer_sharing_nodate}. Recognizing this, our pilot testing revealed the potential to enhance interaction with improved situation awareness, thus refining the dialogue flow.

\begin{figure}
    \centering
    \includegraphics[width=0.95\linewidth]{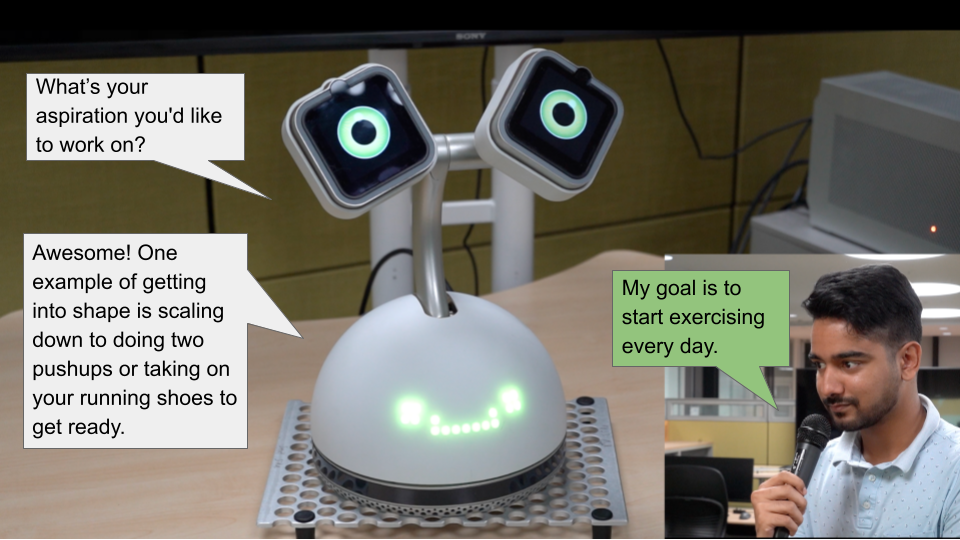}
    
    \caption{Illustration of the Tiny Habit Dialogue} 
    \Description{This figure shows a person interacting with the robot Haru. There is a speech bubble above Haru with the text, 'What’s your aspiration you'd like to work on?' The person responds in another bubble: 'My goal is to start exercising every day.' Haru replies with, 'Awesome! One example of getting into shape is to scale down to doing two pushups or putting on your running shoes to get ready.'}
    \label{fig:enter-label}
    \vspace{-5mm} 
\end{figure}

\begin{itemize}
    \item Adaptive Responses and Dialogue Branching
\end{itemize}
Haru was programmed to tailor its responses based on the user's current environment and expressed preferences. For instance, if the user expressed a desire to skip a story, Haru responded with, ``Let's get right to the point then."
Tailored suggestions were also provided in various parts of the dialogues, depending on the user's replies. For example, when the user was searching for the right moment for their new routine, Haru would provide options such as ``After I pour my cereal'' for mornings, ``After I brush my teeth'' for evenings, and ``After I log into my computer'' for work settings.

\begin{itemize}
    \item Branching Based on User Input
\end{itemize}
The dialogue was scripted with branches to react more effectively to user inputs. For example, when Haru asks, ``Do you like science?'' and the user replies, ``Yes, it is cool,'' Haru responds enthusiastically. If the reply is ``I think it is boring,'' Haru adapts with a more engaging response, ``Oh, so you think it is boring? Let me tell you something."

\begin{itemize}
    \item Personalization and Entity Capture Override
\end{itemize}
To provide personalized coaching, Haru utilized captured entities throughout the interaction. This included using participant's names in dialogues (e.g., ``Oh, \{name\}, what a ride we had'') and incorporating them into the context of the dialogue (e.g., ``Your anchor, along with the tiny habit, might sound something like - 'After I finish my sandwich, \{Tiny Habit of the person\}'").

Recognizing the need to capture entities while also evaluating whether a question required a yes/no response, we implemented an override system. This allowed Haru to capture entities effectively while still enabling users to ask for help or clarification.

\vspace*{-2mm}
\subsection{Error Mitigation}
During pilot testing, we identified the need for a robust help system. This led to the development of a repeat function, enabling Haru to reiterate the last dialogue section upon request (e.g., ``Can you repeat that?"). Additionally, users were given the option to seek help (e.g., ``I don't know, can you please help?"), utilizing the dialogue system's intent recognition feature. Haru also acknowledges and corrects misunderstandings or errors, making statements like, ``I'm sorry. I didn't catch that.'' or ``It was tricky. I couldn't record your anchor moment.'', using apology as a possible trust repair strategy \cite{esterwood_having_2022,jelínek2023trust}.  Such features, including a repeat system, enhance the user experience by providing support and fostering a sense of understanding, which is crucial in effective human-robot interactions.

\vspace*{-2mm}
\subsection{Procedure}

To assess the effectiveness of our interventions, we conducted an experiment at Honda Research Institute in Japan. Participants included employees and interns proficient in English, chosen through convenience sampling. This approach allowed us to compare results with a prior study \cite{jelinek_role_2021}. Participants were first provided with a written consent form, ensuring voluntary participation and agreement with data handling. They were then familiarized with the social robot and the behavior change session, with an opportunity to ask questions before starting.

The sessions with a physically embodied Haru, lasting approximately 12 minutes, were recorded using dual cameras: one capturing participant reactions and the other focusing on the robot. Post-session, participants completed a questionnaire to evaluate the session quality. This included selected items from the Godspeed questionnaire \cite{bartneck_measurement_2008} focusing on anthropomorphism, animacy, and likeability, alongside the MDMT trust scale \cite{ullman_mdmt_2020}, and a set of open questions. We also gathered basic demographic data and inquired about prior experiences with robots.

\balance
\vspace*{-2mm}
\section{Results}
A group of 12 individuals with an average age of 29 years (Stdev=5.72 years) participated in the experiment.
Most participants were male, constituting 83\% of the sample, with 17\% identifying as female. Regarding previous robot experience, 25\% of participants have only seen robots a few times in reality, while the remaining 75\% have had a more direct engagement, either playing or regularly working with robots. The participants' origins were predominantly Asian (58\%), followed by North Americans (25\%) and Europeans (17\%).

In data analysis, conducted using SPSS, we compared our results with the data collected in Condition 1 of the previous experiment \cite{jelinek_role_2021}. Utilizing an Independent Samples T-Test, our findings supported the hypothesis that the changes made in our experiment had a significant effect on the session outcomes.

In the newly developed dialogue, Haru was perceived more positively compared to earlier versions. The statistical analysis revealed that Haru was perceived as significantly less fake (p = 0.042), significantly more lively (p = 0.002), and interactive (p = 0.005). These results indicate a notable shift in participants' perceptions, aligning with the objectives of our dialogue enhancements.

The qualitative data from the experiment with Haru revealed a mix of positive and negative reactions. Participants generally appreciated Haru's expressive characteristics, such as its positive demeanor, movements, and specific routines like chuckling, which added to its human-like appeal. The flow of conversation with Haru was perceived as natural, with an ability to maintain context, and its responsiveness and timing were praised. Visual elements like eye animations and movements also enhanced the experience. Overall, the interaction was described as enjoyable. However, issues with the ASR system were noted multiple times (including cases where the robot 'misunderstood' users and followed incorrect dialogue branches), which likely impacted the overall session evaluation. 

See the \href{https://youtu.be/vayWxkbnz78}{experiment video} for implementation examples: \href{https://youtu.be/vayWxkbnz78}{link}.

\begin{table}[ht]
\centering
\begin{tabular}{lccc}
\hline
Item & MD & p & SD \\
\hline
\textbf{Anthropomorphism} & & & \\
Fake/Neutral & -0.612 & 0.042 & 0.237 \\
Machinelike/Humanlike & -0.100 & 0.282 & 0.288 \\
Unconscious/Conscious & 0.067 & 0.307 & 0.315 \\
Artificial/Lifelike & -0.017 & 0.337 & 0.310 \\
Moving rigidly/elegantly & 0.092 & 0.267 & 0.309 \\
\textbf{Animacy} & & & \\
Dead/Alive & -0.083 & 0.293 & 0.292 \\
Stagnant/Lively & -1.117 & 0.002 & 0.244 \\
Mechanical/Organic & -0.117 & 0.265 & 0.267 \\
Artificial/Lifelike & -0.083 & 0.298 & 0.321 \\
Inert/Interactive & -0.850 & 0.005 & 0.214 \\
Apathetic/Responsive & -0.233 & 0.192 & 0.270 \\
\textbf{Likability} & & & \\
Dislike/Like & -0.450 & 0.088 & 0.267 \\
Unfriendly/Friendly & -0.250 & 0.159 & 0.230 \\
Unkind/Kind & -0.250 & 0.139 & 0.203 \\
Unpleasant/Pleasant & -0.267 & 0.146 & 0.226 \\
Awful/Nice & -0.133 & 0.230 & 0.209 \\
Incompetent/Competent & 0.405 & 0.031 & 0.208 \\
\hline
\end{tabular}
\caption{Summary of mean differences and statistical significance for the measured items.}
\label{tab:mean_differences}
\vspace{-7mm} 
\end{table}

\vspace*{-2mm}
\section{Conclusion}
Our research presents an advancement in the field of human-robot dialogue design, especially in the context of autonomous behavior change coaching with Haru. Through rigorous pilot testing and evaluation, we developed a refined approach to dialogue design, enhancing situation awareness and trust-building strategies. The implementation of a fully autonomous dialogue system, enriched with nuanced situation awareness, robot emotiveness, and a variety of trust-building strategies, marked a novel approach in this domain. The effectiveness of these interventions was reflected in our statistical analysis, showcasing the positive impacts of these innovations. However, the study also revealed potential flaws, such as issues with the Automatic Speech Recognition (ASR) system, which provide valuable insights for future improvements.

Our work not only contributes to the general knowledge of Haru dialogue design but sets a precedent for future research and applications in robot-mediated behavior coaching, highlighting the potential of empathetic, context-aware, and trustworthy interactions in this evolving field. Furthermore, the knowledge and insights gained from our Design Thinking (DT) approach provide a robust foundation for designing future conversations not only for Haru, showcasing how nuanced and effective dialogue design can significantly enhance user experience in diverse contexts.

\vspace*{-2mm}
\section{Future work}

For future work, we aim to utilize the developed dialogue as a dynamic platform for continued research, investigating the specific effects of the isolated dialogue improvements on coaching efficacy and trust. Additionally, we might focus on long-term evaluations to understand Haru's sustained impact in behavior change coaching. Addressing technical challenges and further development of the dialogue system refinement might also be future research.

\section{ACKNOWLEDGMENT}

This project was funded by Honda Research Institute Japan and the Independent Research Fund Denmark, grant number 1032-00311B. 
\bibliographystyle{plain}
\bibliography{mylib}

\end{document}